\def\ARXIVVERSION{1}
\documentclass[10pt,journal,compsoc]{IEEEtran}

\ifCLASSOPTIONcompsoc
  \usepackage[nocompress]{cite}
\else
  \usepackage{cite}
\fi
\usepackage{amsmath}
\usepackage{amssymb}
\usepackage{amsfonts}
\usepackage{array}
\usepackage{booktabs}
\usepackage{capt-of}
\usepackage{graphicx}
\usepackage{makecell}
\usepackage{multirow}
\usepackage{tabularx}
\usepackage[table]{xcolor}
\usepackage{tikz}
\usepackage{url}
\usepackage{algorithm}
\usepackage{algpseudocode}
\usepackage[hidelinks]{hyperref}
\usepackage{microtype}
\usepackage{balance}
\usetikzlibrary{arrows.meta,positioning,calc,fit,backgrounds,shapes.geometric,shapes.misc}

\graphicspath{{figures/}}
\definecolor{MAABlue}{HTML}{0877B9}
\definecolor{MAAGreen}{HTML}{087F5B}
\definecolor{MAARed}{HTML}{B83A2E}
\definecolor{MAALight}{HTML}{EAF4FA}
\newcommand{\best}[1]{\textcolor{MAAGreen}{\textbf{#1}}}
\newcommand{\loss}[1]{\textcolor{MAARed}{#1}}
\newcommand{\MAA}{\textsc{MAA}}

\newenvironment{fixedfigure}{\par\medskip\noindent\begin{minipage}{\columnwidth}\centering}{\end{minipage}\par\medskip}
\newenvironment{fixedtable}{\par\medskip\noindent\begin{minipage}{\columnwidth}\centering}{\end{minipage}\par\medskip}
\renewcommand{\arraystretch}{1.12}
\begin{document}
\def\floatpagepagefraction{0.7}
\def\textpagefraction{.01}

\title{What Softmax Throws Away: Mass-Aware Attention for Evidence Accumulation}

\author{Minwoo~Yu and Young-guk~Ha%
\IEEEcompsocitemizethanks{%
\IEEEcompsocthanksitem M. Yu and Y.-G. Ha are with the Smart Computing Laboratory, Department of Computer Science \& Engineering, Konkuk University, Seoul 05029, Republic of Korea.
E-mail: \{snowypainter, ygha\}@konkuk.ac.kr.
\IEEEcompsocthanksitem Y.-G. Ha is the corresponding author.}}

\ifdefined\ARXIVVERSION
  \markboth{}{}
\else
  \markboth{IEEE Transactions on Pattern Analysis and Machine Intelligence}%
  {Yu and Ha: What Softmax Throws Away}
\fi

\IEEEtitleabstractindextext{%
\begin{abstract}
High task performance alone does not establish whether a model retains the structural information it uses for prediction in its internal representation. Temporal graph models, in particular, can achieve high future-link AUC while basic graph statistics remain inconsistently recoverable from the same representation. We identify one structural source of this gap in the weighted averaging performed by standard attention. When an evidence pattern is repeated, the numerator and denominator of standard attention grow at the same rate; consequently, inputs with different amounts of accumulated evidence can produce the same aggregate.

We address this limitation with \textbf{Mass-Aware Attention (MAA)}, which generalizes standard $L_1$ normalization to an $L_p$ family of operators. By making the numerator and denominator scale at different rates under repetition, MAA retains the effective number of contributing inputs in the magnitude of the representation. It requires no additional supervision, parameters, hidden dimensions, or explicit count features, and recovers standard attention as the special case $p=1$.

Across four CTDG models and three datasets, MAA improves future-link AUC in 11 of 12 model--dataset cells. Linear recovery from the same hidden representation increases by 4.49\% on average, while preferential-attachment recovery improves in all 12 cells and remains significant after family-wise correction. We confirm this effect in CTDG and observe consistent cross-domain evidence in four additional settings: candidate-mark counts in MTPP, pair recurrence in TKG, retained-passage conditions in RAG, and local event counts in STPP. Information accessibility and task utility remain distinct: NLL improves in MTPP and ranking is largely preserved in TKG, whereas the additional information in RAG does not improve the diagnostic head, and a downstream LayerNorm can erase the signal in STPP. These results establish MAA as a general normalization principle for improving the informativeness of predictor-facing representations by controlling the repetition invariance hard-coded into standard attention.
\end{abstract}

}

\maketitle
\IEEEdisplaynontitleabstractindextext
\IEEEpeerreviewmaketitle

\section{Introduction}
\label{sec:intro}

High benchmark performance shows how accurately a model distinguishes future links, but it does not reveal how much pre-query graph structure---such as pair frequency or degree accumulation---remains in the predictor-facing representation. This distinction is especially important for temporal graphs, where repeated interactions, node activity, recency, and periodicity can all support future-link prediction. Yet simple heuristics can achieve strong benchmark performance because of easy negative sampling and recurring edges \cite{poursafaei2022better,cornell2025heuristics}, and model predictions do not consistently reflect several graph characteristics \cite{hayes2025what}. A model that learns a temporal graph effectively should therefore make prediction-relevant graph statistics accessible in the representation supplied to its predictor. In this work, statistic recovery is not a substitute for task performance; it is a \emph{representation audit} of this condition.

We study a small but pervasive blind spot shared by attention models across domains. Standard softmax attention returns a weighted average of its input values. When the same evidence pattern is repeated, its numerator and $L_1$ denominator can grow together, leaving the aggregate unchanged. The operator preserves which evidence is relatively important while weakening how broadly that evidence has accumulated. This effect can appear as low graph-statistic recovery in CTDG, weak accessibility of cumulative candidate-mark occurrences in MTPP, query-specific pair recurrence in TKG, evidence completeness across passages in RAG, and event counts around a candidate location in STPP.

We introduce \textbf{Mass-Aware Attention (MAA)} to address this blind spot. The only modification is to generalize the $L_1$ norm in the attention denominator to an $L_p$ norm. Standard attention becomes the point $p=1$; for $p>1$, the numerator and denominator scale at different rates under repetition. MAA thereby preserves the weighted direction while retaining an input-dependent scale derived from the effective support of the attention distribution. It uses no explicit count, auxiliary target, additional parameter, or hidden coordinate. Instead, it exposes the exact repetition invariance hard-coded by standard attention as a continuous control axis.

\begin{figure*}[!t]
\centering
\includegraphics[width=\textwidth]{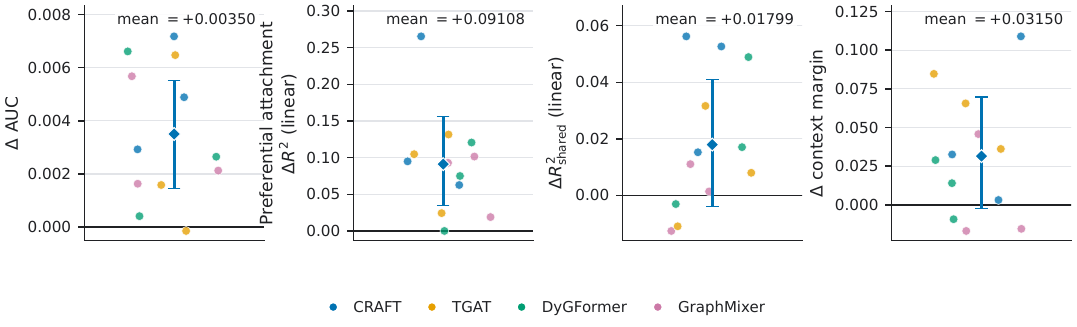}
\caption{Main CTDG results. Linear recovery and context sensitivity increase on average. Confirmatory evidence is strongest for preferential-attachment recovery, which improves in 12/12 cells, and future-link AUC, which improves in 11/12 cells. Effect sizes vary across probes and statistics.}
\label{fig:main-results}
\end{figure*}

Our evaluation follows a hierarchy of evidence. The primary question is whether target information becomes more recoverable from the same predictor-facing hidden representation $Z$. We then measure whether this change preserves or improves the original task. Finally, a $2\times2$ representation audit, an explicit-cardinality control, output normalization, and power sensitivity distinguish MAA from coordinate exposure and uniform changes in representation norm. Figure~\ref{fig:main-results} summarizes the central CTDG results, and Table~\ref{tab:domain-scope} shows how the common question is instantiated in each domain.

\begin{table*}[!t]
\centering
\caption{Evidence-accumulation problem and evaluation target in each domain. Every recovery result is measured from an equal-dimensional predictor-facing representation $Z$ without auxiliary coordinates.}
\label{tab:domain-scope}
\small
\setlength{\tabcolsep}{4.5pt}
\begin{tabularx}{\textwidth}{@{}l X X X@{}}
\toprule
Domain & Input distinction & Information recovered from $Z$ & Task-utility metric \\
\midrule
CTDG & Differences in repeated interactions and temporal neighborhoods
& Pair frequency, recency, persistence, periodicity, granularity, density, homophily, and preferential attachment
& Future-link ROC--AUC and context sensitivity \\
MTPP & Accumulation of preceding events supporting a candidate mark
& Global and recent occurrence counts of the candidate mark
& Point-process/mark NLL, mark accuracy, and time error \\
TKG & Repeated historical facts connecting the current head and candidate tail
& Pair recurrence, relation diversity, and two-hop path count
& Filtered MRR, Hits@10, and hard-negative AUC \\
RAG & Complete supporting passages versus inputs missing part of the evidence
& Retained-passage condition with literal-duplicate cues removed
& Diagnostic AUC for complete versus partial evidence \\
STPP & Preceding events accumulated around a candidate location
& Local event counts at multiple radii
& NELBO and spatial/temporal log-likelihood \\
\bottomrule
\end{tabularx}
\end{table*}

Our contributions are as follows.
\begin{itemize}
    \item We reinterpret the repetition invariance of standard attention as the $p=1$ special case of an $L_p$ normalization family and introduce \MAA, a parameter-free mechanism for controlling it.
    \item We jointly evaluate recovery from the same representation, behavioral sensitivity, and task performance across four CTDG architectures, three datasets, and three seeds. Preferential-attachment recovery improves in all 12 model--dataset cells, and AUC improves in 11.
    \item We extend the evaluation to MTPP, TKG, RAG, and STPP, showing that one normalization principle improves access to distinct forms of accumulated evidence.
    \item We characterize the operational boundary of \MAA through information that becomes accessible but remains unused in RAG and information that a downstream normalization removes in STPP.
\end{itemize}

\section{Related Work}
\label{sec:related}

This section reviews temporal representation auditing and attention normalization, and distinguishes \MAA from explicit cardinality injection.

Standard scaled dot-product attention forms a weighted average of values using normalized coefficients \cite{vaswani2017attention}. We study the repetition invariance imposed by the denominator of this basic operator.

\subsection{Temporal Graph Learning and Representation Auditing}
TGAT constructs inductive node representations using time encoding and temporal attention \cite{xu2020tgat}. GraphMixer combines link and node encoders with an MLP mixer \cite{cong2023graphmixer}, while DyGFormer encodes interaction sequences as patches \cite{yu2023dygformer}. CRAFT achieves strong future-link performance using node identities and target-aware cross-attention \cite{yi2025craft}. \MAA does not replace these backbones; it modifies only the normalization at the attention boundary that forms the predictor-facing representation.

In dynamic link prediction, recurring edges and memorization can dominate under particular evaluation protocols \cite{poursafaei2022better}. Hayes et al.\ show through interventions that model predictions reflect temporal graph characteristics unevenly \cite{hayes2025what}. Their analysis is behavioral, whereas ours examines frozen internal representations. We measure the recoverability of query-level graph statistics and test an operator that improves their accessibility without statistic supervision.

\subsection{Representation Probing}
Linear probes are a standard tool for measuring whether a target is accessible through a simple readout from a frozen representation \cite{alain2017understanding,conneau2018cram,hewitt2019designing}. Because a probe may itself learn the target or exploit surface correlations, its capacity and held-out protocol must be interpreted separately \cite{kunz2020classifier}. We report Ridge and ExtraTrees probes on chronological held-out splits, never propagate probe gradients into the original model, and evaluate recovery together with behavioral sensitivity and task performance. This design separates information that is readable from information that the predictor actually uses.

\subsection{Cardinality Preservation and Probability Normalization}
Cardinality-Preserved Attention (CPA) identifies the inability of graph attention to distinguish multisets with different multiplicities and augments the aggregate with an unweighted sum or explicit neighborhood cardinality \cite{zhang2020cardinality}. Its motivation is closely related, but its information source and operator differ from ours. CPA injects the observed token count or an unweighted aggregate through a separate branch. \MAA instead derives its scale from the effective support and concentration of the attention distribution already computed by the model. It can therefore assign different scales to inputs with identical cardinality but different attention distributions, while $p$ continuously controls the strength of the invariance.

Normalized Attention Without Probability Cage replaces softmax with normalization based on the mean and variance of logits and learnable gain and bias, motivated by the probability-simplex and convex-hull constraints \cite{richter2020probability}. Affine-Scaled Attention learns an input-dependent affine transformation after softmax \cite{bae2026affine}. In contrast, \MAA preserves the standard weighted direction and learns no additional gate: its scale is deterministically computed from the normalized attention $\alpha$. The distinctive contribution of \MAA is thus to expose the $L_1$ repetition invariance hard-coded into standard attention as an explicit control axis in an $L_p$ family.

\subsection{Evidence Accumulation Across Domains}
In temporal point processes, event history determines mark-specific conditional intensities and the distribution of the next event, which recurrent and attention-based models represent in different ways \cite{du2016recurrent,mei2017neural,zuo2020transformer,yang2022transformer}. In temporal knowledge graph forecasting, recurrent facts and relational paths support candidate-entity ranking \cite{jin2020recurrent,han2021explainable,zhu2021cygnet}. Multi-hop QA requires integrating supporting facts distributed across passages \cite{yang2018hotpotqa,izacard2021leveraging}, while spatio-temporal point processes condition spatial intensity on local event density and revisit patterns \cite{chen2021neuralstpp,zhou2022deepstpp}. Although these tasks and data structures differ, they share an attention boundary that conveys accumulated evidence to a predictor through a weighted average. We therefore use domain-specific targets that directly represent accumulation rather than imposing one statistic across domains.

Layer normalization standardizes hidden-state scale \cite{ba2016layer}, and its placement before or after attention changes optimization and information flow in Transformers \cite{xiong2020layer}. Our STPP analysis identifies a concrete boundary at which this general architectural choice determines whether the input-dependent amplitude introduced by \MAA reaches the predictor-facing representation.

\section{Mass-Aware Attention}
\label{sec:method}

This section formalizes the repetition invariance of standard weighted averaging and presents the $L_p$ normalization that relaxes it continuously.

\subsection{Repetition Invariance of Standard Attention}
Let the inputs participating in the aggregation for one query be $X=\{(m_i,v_i)\}_{i=1}^{n}$, where $v_i$ is a value and $m_i=\exp(a_i-\max_j a_j)$ is a numerically stabilized positive attention weight. Under this notation, standard softmax aggregation is defined by
\begin{equation}
A_1(X)=\frac{\sum_i m_i v_i}{\lVert m\rVert_1}.
\label{eq:softmax}
\end{equation}
Equation~\eqref{eq:softmax} divides the weighted sum of values by the total $L_1$ mass, forcing the coefficients to sum exactly to one. To examine how this operator treats multiplicity, let $X^{(r)}$ contain $r$ copies of the same weighted-value motif. Replication multiplies both the numerator and denominator by $r$, yielding
\begin{equation}
A_1(X^{(r)})=A_1(X).
\label{eq:collision}
\end{equation}
Equation~\eqref{eq:collision} is a mathematical thought experiment rather than a training procedure that duplicates real observations. It demonstrates that standard weighted averaging preserves the relative evidence pattern but may map different numbers of repetitions of that pattern to the same aggregate.

\begin{figure*}[!t]
\centering
\includegraphics[width=\textwidth]{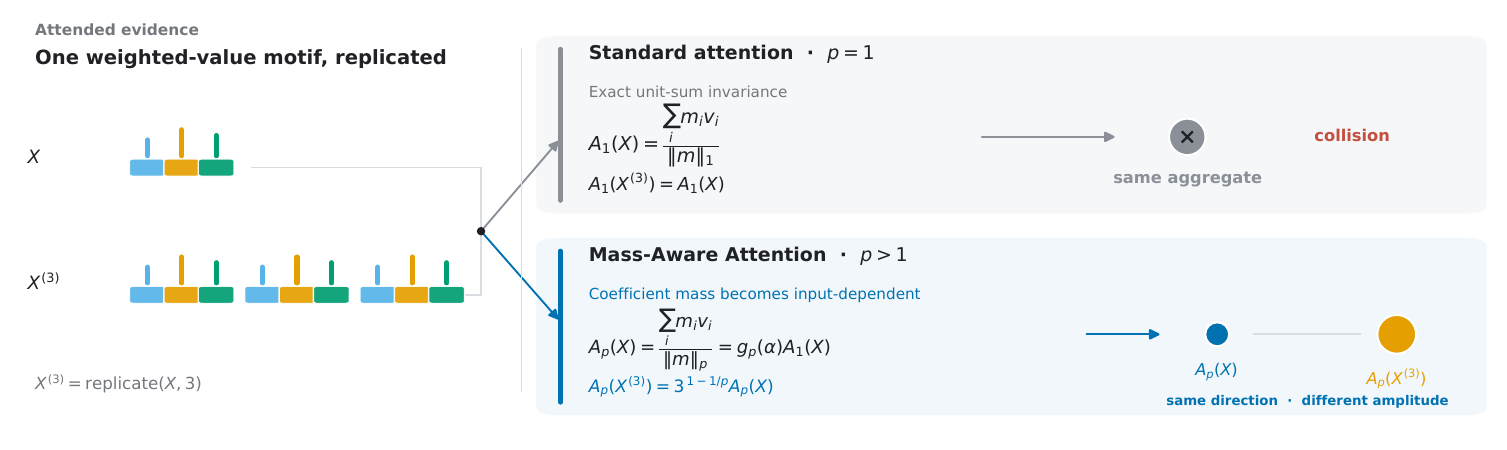}
\caption{Intuition behind Mass-Aware Attention. Standard attention can map repetitions of the same weighted-value pattern to an identical aggregate. \MAA makes the numerator and denominator grow at different rates under repetition, preserving the weighted direction while retaining an effective-input-dependent amplitude.}
\label{fig:method}
\end{figure*}

Figure~\ref{fig:method} visualizes this invariance and the change introduced by \MAA. Standard attention collapses replicated inputs to the same aggregate, whereas \MAA preserves the weighted direction while retaining an amplitude difference induced by repetition.

\subsection{The $L_p$ Normalization Family}
\MAA extends the exact $L_1$ normalization hard-coded into standard attention to a continuous family of operators. It replaces only the denominator in~\eqref{eq:softmax} with an $L_p$ norm:
\begin{equation}
A_p(X)=\frac{\sum_i m_i v_i}{\lVert m\rVert_p}, \qquad p\ge 1.
\label{eq:maa}
\end{equation}
Standard attention is the special case $p=1$. The effect of \MAA becomes clearer using normalized attention $\alpha=m/\lVert m\rVert_1$. Rearranging~\eqref{eq:maa} gives
\begin{equation}
A_p(X)
=\underbrace{\frac{1}{\lVert\alpha\rVert_p}}_{g_p(\alpha)}
A_1(X).
\label{eq:gate}
\end{equation}
Equation~\eqref{eq:gate} shows that \MAA preserves the standard weighted direction $A_1(X)$ and retains only an input-dependent scale $g_p(\alpha)$. This scale is computed from the normalized attention distribution itself rather than from an injected token count. For $p>1$, it satisfies
\begin{equation}
1\le g_p(\alpha)\le n^{1-1/p}.
\label{eq:gate-bound}
\end{equation}
The scale approaches one when attention concentrates on a few inputs and grows as attention spreads over more inputs. In terms of R\'enyi entropy $H_p(\alpha)$, it can equivalently be written as $g_p(\alpha)=\exp((1-1/p)H_p(\alpha))$. Thus, $g_p(\alpha)$ is a monotonic transformation of the attention-weighted effective support.

This generalization separates the growth rates of the numerator and denominator under repetition. For $X^{(r)}$, the numerator grows by $r$, while the $L_p$ denominator grows by $r^{1/p}$. The resulting response is
\begin{equation}
A_p(X^{(r)})=r^{1-1/p}A_p(X).
\label{eq:replicate}
\end{equation}
At $p=1$, the two rates cancel exactly and recover~\eqref{eq:collision}. For $p>1$, the effective size of the replicated input remains as a controlled amplitude. Importantly, this is not the arbitrary absolute mass of unnormalized scores. \MAA conveys a reproducible scale derived from the concentration and effective support of normalized attention.

\subsection{Implementation}
\MAA changes only the computation of the normalization denominator in an existing attention pipeline. Algorithm~\ref{alg:maa} summarizes the operation for one query and one attention head.

\begin{algorithm}[t]
\caption{\MAA at One Attention Boundary}
\label{alg:maa}
\begin{algorithmic}[1]
\Require logits $a$, values $V$, valid mask $M$, power $p\ge1$
\State $s\gets\max_{i:M_i=1}a_i$
\State $m_i\gets M_i\exp(a_i-s)$
\State $d_p\gets(\sum_i m_i^p)^{1/p}$
\State $\beta_i\gets m_i/\max(d_p,\epsilon)$
\State \Return $\sum_i\beta_iV_i$
\end{algorithmic}
\end{algorithm}
The operation is applied independently to each query and head, with masked tokens excluded from the norm. Logits, value projections, residual connections, and decoders remain unchanged, so neither parameter count nor hidden dimensionality increases. Time and space complexity remain of the same order as standard attention. This minimal change enables the same principle to be applied across domains, although a strong scale-removing normalization after~\eqref{eq:gate} can erase the signal.

\section{Evaluation Principles and Protocol}
\label{sec:protocol}

The evaluation consistently separates information accessibility in the same representation, original-task utility, and power selection.

\subsection{Representation Recovery}
After training, we freeze the hidden representation $Z$ supplied to the predictor. Probes are trained on a chronological held-out split; neither recovery targets nor probe gradients participate in training the original model. For CTDG, we use both linear Ridge and ExtraTrees. The linear probe measures linear accessibility from the same $Z$, whereas ExtraTrees measures nonlinear accessibility. Unless otherwise stated, the primary cross-domain recovery measure uses a linear probe. Continuous targets are evaluated with held-out $R^2$ and binary conditions with AUC.

\textbf{Shared recovery} directly compares the equal-dimensional hidden representations $Z_B$ and $Z_{\MAA}$ of the baseline and \MAA. \textbf{Beyond-posthoc recovery} supplies both probes with the same type of auxiliary coordinate $c$ computed during attention and measures the remaining difference. The former tests whether the final representation itself changes; the latter tests whether the learned representation changes beyond simply exposing an already computed coordinate. \textbf{Context margin} is the difference in prediction between the original query context and a condition with a substituted source context. A larger margin indicates greater sensitivity of model output to contextual differences.

\subsection{Separating Power Selection from Evaluation}
Because $p$ defines the operator family, we select it through a constrained model-selection problem aligned with the purpose of MAA. The primary objective is information accessibility in the predictor-facing representation, while preservation of original-task utility is a feasibility constraint. The two outcomes are therefore not combined post hoc with equal status. We use a pre-specified lexicographic order: the selection split first identifies powers that preserve baseline task utility, and the pre-specified recovery endpoint selects a power within this feasible set. We then freeze $p$ and evaluate final task performance and recovery once on held-out conditions that did not participate in selection.

For CTDG, we instantiate this rule through leave-one-model-out selection. Cells from three models select $p$, and the excluded model provides evaluation; all four folds select $p=1.1$. Consequently, none of the 12 reported CTDG cells uses a power selected from results of its own model. RAG applies the same task-preserving recovery rule on an independent validation split and reports results on a separate test split. MTPP fixes $p=1.3$ in an earlier pilot and evaluates it on new likelihood and time endpoints. TKG separates an ICEWS14 pilot from ICEWS18 validation, fixing $p=1.2$ and $p=1.02$, respectively, before test evaluation. STPP evaluates $p=1.05$ in a matched pre-norm architecture after diagnosing the post-norm boundary.

\paragraph{Reusable decision rule.}
Algorithm~\ref{alg:power-selection} operationalizes the procedure for a new model or domain. The audit target, candidate grid, and task-utility tolerance are fixed before opening the test set. We use a small grid dense near the standard operator, $\mathcal P=\{1,1.02,1.05,1.1,1.2,1.3\}$. Let $U_{\mathrm{sel}}(p)$ denote the original-task metric oriented so that larger is better, and let $R_{\mathrm{sel}}(p)$ denote recovery from the same predictor-facing $Z$. In CTDG, a candidate is feasible only when its mean selection-fold AUC is at least that of standard attention ($p=1$). In RAG, feasibility analogously requires validation diagnostic AUC at least as high as the $p=1$ baseline. Both studies therefore use $\epsilon_U=0$ and select the power with the highest pre-specified recovery endpoint among feasible candidates. Future applications may instead pre-specify a domain-accepted practical-equivalence margin.

\begin{algorithm}[t]
\caption{Validation-Based Power Selection for \MAA}
\label{alg:power-selection}
\begin{algorithmic}[1]
\Require candidate grid $\mathcal P$, selection split, fixed audit target, task tolerance $\epsilon_U$
\State Evaluate $U_{\mathrm{sel}}(p)$ and $R_{\mathrm{sel}}(p)$ for every $p\in\mathcal P$
\State $\mathcal F\gets\{p:U_{\mathrm{sel}}(p)\ge U_{\mathrm{sel}}(1)-\epsilon_U\}$
\If{$\mathcal F=\varnothing$}
  \State \Return $p^\star=1$
\EndIf
\State $p^\star\gets\arg\max_{p\in\mathcal F}R_{\mathrm{sel}}(p)$
\State Break recovery ties toward the smaller $p$
\State \Return frozen $p^\star$; evaluate the held-out test once
\end{algorithmic}
\end{algorithm}

This lexicographic rule gives the two design goals explicit priority: the task constraint protects predictive utility, and the recovery objective selects the more informative representation within that constraint. Candidate comparison, feasibility filtering, and tie breaking are completed entirely on the selection split, making held-out recovery and task utility outcomes of evaluation rather than inputs to selection.

This tunability is a design advantage of MAA. Standard attention fixes exact $L_1$ normalization, and hence $p=1$, for every input and task. MAA exposes this implicit design decision as an explicit hyperparameter, allowing practitioners to calibrate the strength of repetition invariance through standard validation without observing test outcomes. Because input length, attention concentration, and downstream normalization alter the distribution of $g_p(\alpha)$ induced by the same $p$, an explicit and reproducible selection rule is more appropriate than a universal constant. When no meaningful audit target is available, $p=1.1$, the common CTDG value, can serve as a default, subject to the same validation check on task utility.

\subsection{Statistical Inference}
Seeds within the same model--dataset pair are repeated measurements of one condition, not independent research settings. Confirmatory CTDG inference therefore first averages the three seed-level effects within each of the 12 model--dataset cells. We apply two-sided Wilcoxon signed-rank tests and jointly adjust the six aggregate endpoints with Holm correction. We additionally report 95\% intervals from 20,000 crossed hierarchical bootstrap iterations that resample models and datasets independently and then resample seeds within each selected cell. The 36 seed-level results serve as a sensitivity analysis of direction and variability.

The cross-domain extension is exploratory, and all paired Wilcoxon tests are therefore two-sided. In small samples, conclusions jointly consider effect size, the number of improving repetitions, and original-task performance rather than relying on significance alone.

\section{Core Validation on CTDG}
\label{sec:ctdg}

CTDG serves as the core validation domain because prior work has already identified a concrete gap between benchmark performance and the graph mechanisms captured by temporal models. Dynamic link-prediction results can depend strongly on negative sampling and recurring edges \cite{poursafaei2022better}, and recent evidence shows that model predictions do not consistently reflect several temporal graph characteristics \cite{hayes2025what}. We extend this behavioral question to recovery from predictor-facing representations and conduct our broadest comparison and confirmatory inference in CTDG, using four models and three datasets.

\subsection{Setup and Representation Gap}
\begin{fixedtable}
\centering
\captionof{table}{Experimental setup for the core CTDG study.}
\label{tab:ctdg-setup}
\small
\begin{tabularx}{\columnwidth}{@{}l X@{}}
\toprule
Component & Setting \\
\midrule
Models & CRAFT, TGAT, DyGFormer, GraphMixer \\
Datasets & LastFM, MOOC, Wikipedia \\
Repetitions & Seeds 7, 17, and 29; 36 paired runs \\
Training data & Most recent 32,768 events in chronological order; 70/30 split \\
Objective & Original future-link BPR loss \\
Method & Baseline $p=1$; fixed-power MAA $p=1.1$ \\
Probes & Chronological held-out Ridge and ExtraTrees \\
Task metric & ROC--AUC over positives and sampled negatives \\
Inference unit & 12 model--dataset cells after within-cell seed averaging \\
\bottomrule
\end{tabularx}
\end{fixedtable}

We evaluate CRAFT, TGAT, DyGFormer, and GraphMixer on LastFM, MOOC, and Wikipedia. These temporal interaction benchmarks were released with JODIE \cite{kumar2019predicting}. Every model uses the same recent-event budget and its original future-link objective. The eight recovery targets are pair frequency, recency, persistence, periodicity, granularity, density, homophily, and preferential attachment (PA). PA audits the count-structured information left by repeated edge formation at the node level because it reflects accumulated endpoint degrees \cite{barabasi1999emergence}. Pair frequency directly counts repetitions at the pair level, whereas PA measures degree accumulation at the node level.

Baseline recovery varies substantially across models and statistics. CRAFT is relatively strong on PA and pair frequency, while TGAT recovers density and granularity more accurately. This analysis motivates the central observation that high AUC does not determine the graph-statistic content of a predictor-facing representation.

\subsection{$2\times2$ Representation Audit}
\begin{fixedfigure}
\centering
\includegraphics[width=\columnwidth]{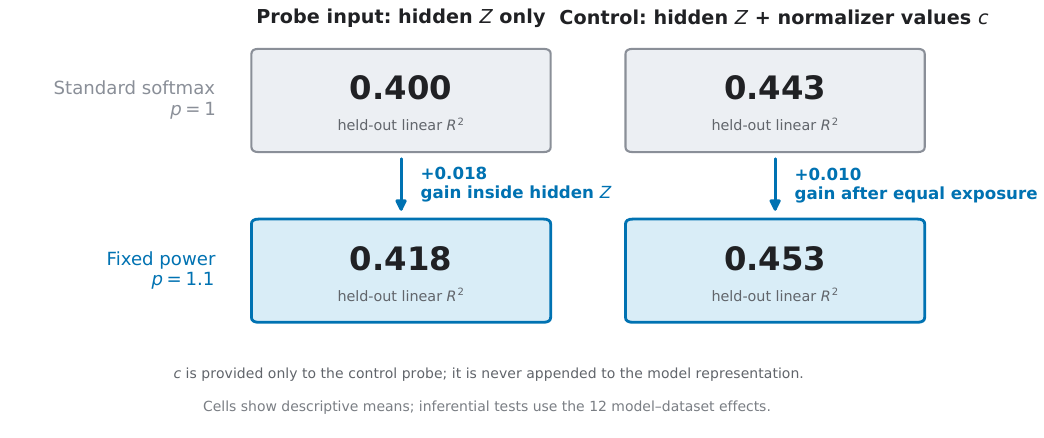}
\captionof{figure}{$2\times2$ audit separating changes in the hidden representation $Z$ learned by \MAA from gains obtained by exposing a normalization-derived coordinate to the probe. The shared comparison directly contrasts $Z_B$ and $Z_{\MAA}$, while the post-hoc comparison supplies the same type of coordinate $c$ to both sides.}
\label{fig:audit}
\end{fixedfigure}
The four conditions in Fig.~\ref{fig:audit} are the baseline representation $Z_B$, its post-hoc augmentation $(Z_B,c_B)$, the equal-dimensional \MAA representation $Z_{\MAA}$, and $(Z_{\MAA},c_{\MAA})$. The primary comparison is $Z_{\MAA}$ versus $Z_B$, which rules out gains caused by providing a new feature to the probe. The beyond-posthoc comparison tests whether a difference remains after exposing an equivalent coordinate on both sides.

\subsection{Aggregate Results}
\begin{fixedtable}
\centering
\captionof{table}{Aggregate CTDG results. Wilcoxon tests use 12 paired effects obtained by first averaging three seeds within each model--dataset cell. Holm correction is applied jointly to the six aggregate endpoints.}
\label{tab:ctdg-main}
\small
\setlength{\tabcolsep}{4pt}
\resizebox{\columnwidth}{!}{\begin{tabular}{@{}lrrrrrr@{}}
\toprule
Measure & Baseline & MAA & Mean change & Improved cells & $p$ & Holm $p$ \\
\midrule
AUC & 0.85985 & 0.86336 & \best{+0.00350 (+0.41\%)} & \best{11/12} & 0.000977 & \best{0.005859} \\
Shared ExtraTrees $R^2$ & 0.52526 & 0.52628 & +0.00102 (+0.19\%) & 4/12 & 0.518555 & 0.518555 \\
Beyond ExtraTrees $R^2$ & 0.55928 & 0.55761 & -0.00168 (-0.30\%) & 5/12 & 0.233398 & 0.466797 \\
Shared linear $R^2$ & 0.40020 & 0.41818 & \best{+0.01799 (+4.49\%)} & 9/12 & 0.034180 & 0.136719 \\
Beyond linear $R^2$ & 0.44278 & 0.45298 & +0.01020 (+2.30\%) & 9/12 & 0.052246 & 0.156738 \\
Context margin & 0.54111 & 0.57261 & \best{+0.03150 (+5.82\%)} & 9/12 & 0.026855 & 0.134277 \\
\bottomrule
\end{tabular}}
\end{fixedtable}

Table~\ref{tab:ctdg-main} shows that AUC rises by 0.41\% on average and improves in 11 of 12 cells. The 12-cell Wilcoxon test, Holm correction, and crossed-bootstrap interval agree on the direction. Shared linear recovery increases by 4.49\% and context margin by 5.82\%. Both effects are positive in 9/12 cells, although uncertainty remains after aggregate family-wise correction. Mean ExtraTrees recovery is nearly unchanged. The clearest representational effect of \MAA is therefore improved linear accessibility from the same $Z$, rather than a universal increase in nonlinear recoverability.

\begin{fixedtable}
\centering
\captionof{table}{Key statistic-wise changes in CTDG. Holm correction over the eight statistics is performed separately for each probe.}
\label{tab:ctdg-statistics}
\scriptsize
\setlength{\tabcolsep}{3pt}
\begin{tabular}{@{}llrrr@{}}
\toprule
Probe & Statistic & $\Delta R^2$ & Cell & Holm $p$ \\
\midrule
Linear & PA & \best{+0.09108} & \best{12/12} & \best{0.003906} \\
Linear & Pair frequency & +0.01392 & 8/12 & 0.187988 \\
Linear & Density & +0.03754 & 9/12 & 0.462891 \\
ExtraTrees & Density & \loss{-0.02170} & 2/12 & \loss{0.039062} \\
\bottomrule
\end{tabular}
\end{fixedtable}

Statistic-level analysis sharpens this result. Linear PA recovery increases by $0.09108$ on average, improves in all 12 cells, and survives Holm correction across the eight statistics (Table~\ref{tab:ctdg-statistics}). Pair frequency has a positive mean effect but remains uncertain after correction, whereas ExtraTrees density decreases significantly. Figure~\ref{fig:statistics} shows both statistic-level uncertainty and model heterogeneity. The most stable gains occur along linear directions tied directly to repetition and degree accumulation, while some nonlinear structure is redistributed.

\begin{fixedfigure}
\centering
\includegraphics[width=\columnwidth]{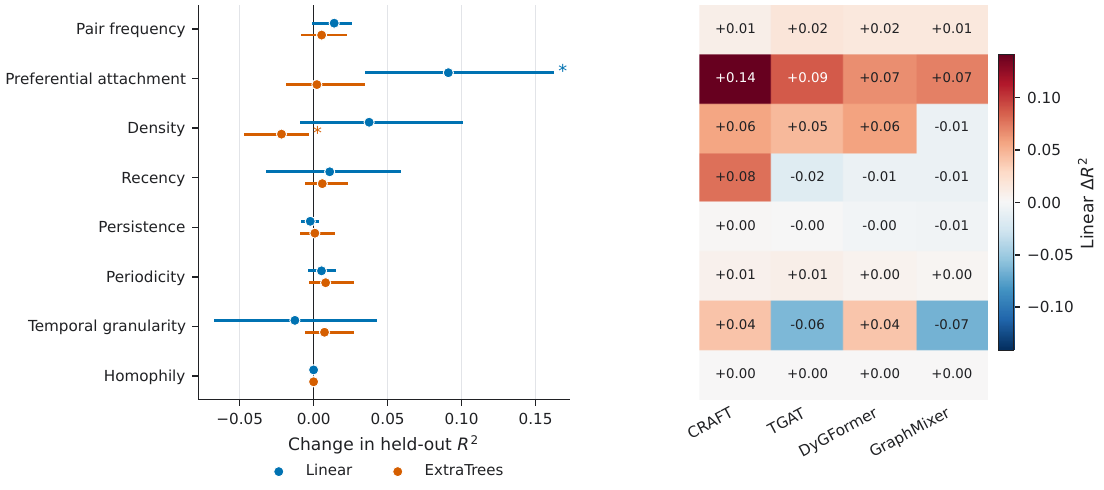}
\captionof{figure}{Statistic-wise changes in recovery and their boundary. Linear recovery improves most strongly and consistently for preferential attachment, with pair frequency changing in the same direction. The decrease in ExtraTrees density shows that accessibility varies across probes and statistics.}
\label{fig:statistics}
\end{fixedfigure}

\subsection{Behavioral Sensitivity}
Context margin measures the score difference between the observed source temporal neighborhood and a substituted neighborhood. Its average increase of 5.82\% shows that the additional probe-accessible information coincides with greater sensitivity of predictive behavior to contextual differences. Because the margin alone does not establish directionally correct causal use, we interpret it together with original-task AUC and statistic interventions.

\section{Cross-Domain Generalization}
\label{sec:cross}

\begin{figure*}[!t]
\centering
\includegraphics[width=\textwidth]{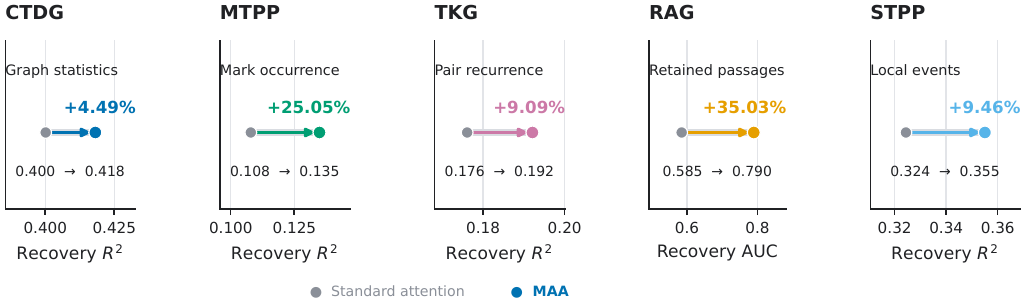}
\caption{Changes in predictor-facing representation recovery across domains. Each value is an equally weighted average over the model--dataset conditions evaluated in that domain. Targets and metrics follow the predictive structure of each domain; the common question is whether evidence-accumulation information becomes more accessible from the same $Z$, without auxiliary coordinates.}
\label{fig:cross-recovery}
\end{figure*}

We extend the CTDG question to four additional domains. For each domain, we first define a count- or support-based statistic that \MAA is expected to retain, and then separately measure its recovery from the same $Z$, the model's behavioral response to evidence manipulation, and utility on the original task. Figure~\ref{fig:cross-recovery} and Table~\ref{tab:cross-summary} place the results from all domains in a common framework.

\begin{fixedtable}
\centering
\captionof{table}{Key cross-domain results. Recovery is the primary endpoint, and task metrics jointly report predictive utility.}
\label{tab:cross-summary}
\small
\setlength{\tabcolsep}{2.5pt}
\begin{tabularx}{\columnwidth}{@{}l X X@{}}
\toprule
Domain & Representation improvement & Task utility \\
\midrule
CTDG & Linear $R^2$ \best{+4.49\%}; PA \best{12/12}
& Link AUC \best{+0.41\%} \\
MTPP & Mark-count $R^2$ \best{+25.05\%}
& Point-process NLL \best{2.59\% lower} \\
TKG & Pair recurrence \best{+9.09\%}
& Hits@10 \best{+1.07\%} \\
RAG & Retained-passage AUC \best{+35.03\%}
& Diagnostic AUC -0.73\% (n.s.) \\
STPP & Local-count $R^2$ \best{+9.46\%}
& Matched NELBO \best{4.46\% better} \\
\bottomrule
\end{tabularx}
\end{fixedtable}

\subsection{MTPP: Accumulated Occurrences of a Candidate Mark}
AttNHP and THP encode event sequences with self-attention to predict the mark and time of the next event \cite{yang2022transformer,zuo2020transformer}. The candidate mark of the next event is conditionally related to the occurrence pattern of the same mark in preceding events. We use candidate-mark counts over the full past and the most recent 32 and 8 events as recovery targets. These quantities directly summarize inputs to the mark-specific conditional intensity. \MAA improves full-history count $R^2$ in all six aggregated model--dataset--seed conditions, yielding a mean improvement of 25.05\%. Concurrently, next-mark NLL decreases by 1.29\% and point-process NLL by 2.59\%. For AttNHP, next-time MAE and RMSE also decrease by 2.14\% and 3.96\%, respectively. The joint improvement in information accessibility and likelihood indicates that the information attenuated by standard weighted averaging has predictive value for this task.

\subsection{TKG: Query-Specific Pair Recurrence}
xERTE expands a query-relevant temporal subgraph through attention to rank future entities \cite{han2021explainable}; this architecture differs from RE-NET, which uses a recurrent event encoder \cite{jin2020recurrent}. For a TKG query $(head, relation, ?, time)$, the number of previous connections between the same head and candidate tail provides more query-specific evidence than global tail popularity. On ICEWS14 and ICEWS18, we recover pair recurrence, the number of relation types observed for the same pair, and the number of historical two-hop paths. The stand-alone diagnostic AUC of pair recurrence exceeds that of global tail count on both datasets, establishing its relevance to the ranking problem.

\begin{fixedtable}
\centering
\captionof{table}{Representative MTPP and TKG results. A downward arrow marks a lower-is-better metric.}
\label{tab:mtpp-tkg}
\small
\resizebox{\columnwidth}{!}{\begin{tabular}{@{}lllrrrr@{}}
\toprule
Domain & Setting & Endpoint & $p=1$ & MAA & Relative change & Improved runs \\
\midrule
MTPP & All 6 settings & Candidate-mark count $R^2$ & 0.10792 & 0.13495 & \best{+25.05\%} & \best{6/6} \\
 &  & Point-process NLL $\downarrow$ & 2.96749 & 2.89063 & \best{-2.59\%} & 5/6 \\
 &  & Next-mark NLL $\downarrow$ & 1.92280 & 1.89791 & \best{-1.29\%} & \best{6/6} \\
\midrule
TKG & ICEWS14 & Pair-recurrence $R^2$ & 0.22149 & 0.24773 & \best{+11.85\%} & \best{3/3} \\
 &  & Hits@10 & 0.68300 & 0.69617 & \best{+1.93\%} & \best{3/3} \\
 & ICEWS18 & Pair-recurrence $R^2$ & 0.13071 & 0.13650 & \best{+4.43\%} & \best{3/3} \\
 &  & MRR & 0.22807 & 0.22760 & -0.20\% & 1/3 \\
\bottomrule
\end{tabular}}
\end{fixedtable}

As shown in Table~\ref{tab:mtpp-tkg}, pair-recurrence recovery increases by 11.85\% on ICEWS14 and by 4.43\% in the deterministic ICEWS18 rerun, with the same direction in all three seeds for both datasets. Hits@10 improves by 1.93\% on ICEWS14. On ICEWS18, MRR decreases by 0.20\% and Hits@10 by 0.43\%, preserving nearly equivalent ranking performance. These results show that \MAA can retain query-specific recurrence more clearly while leaving the ranking function largely intact, even when every task metric does not improve simultaneously.

\subsection{RAG: Separating Accessibility from Utilization}
In multi-hop question answering tasks such as HotpotQA, success requires combining facts distributed across distinct supporting passages rather than merely repeating one passage \cite{yang2018hotpotqa}. FiD encodes retrieved passages independently and combines them through decoder attention \cite{izacard2021leveraging}. We therefore define a retained-passage condition that distinguishes complete evidence from inputs missing part of the supporting passages. Our primary analysis masks duplicate-token and duplicate-passage indicators to prevent literal duplication from becoming a count shortcut. Recovery AUC measures how well this binary condition can be separated from the same $Z$, whereas task AUC measures whether the original diagnostic head assigns higher scores to complete evidence.

\begin{fixedtable}
\centering
\captionof{table}{Boundary results for RAG and STPP. Arrows denote changes from baseline to MAA.}
\label{tab:rag-stpp}
\small
\setlength{\tabcolsep}{3pt}
\begin{tabularx}{\columnwidth}{@{}lXrr@{}}
\toprule
Domain & Endpoint & Baseline $\rightarrow$ MAA & Change \\
\midrule
RAG & Retained-passage recovery AUC & $0.585\rightarrow0.790$ & \best{+35.03\%} \\
 & Diagnostic task AUC & $0.614\rightarrow0.609$ & -0.73\% \\
\midrule
STPP & Predictor-facing count $R^2$ & $0.324\rightarrow0.355$ & \best{+9.46\%} \\
 & Encoder count $R^2$ & $0.452\rightarrow0.493$ & \best{+9.18\%} \\
 & NELBO $\downarrow$ & $-1.675\rightarrow-1.750$ & \best{4.46\% better} \\
\bottomrule
\end{tabularx}
\end{fixedtable}

\MAA raises duplicate-masked retained-passage recovery AUC from 0.58482 to 0.78967 (35.03\%, 5/5). Diagnostic task AUC decreases by 0.73\%, with a two-sided $p=1.0$. As Fig.~\ref{fig:cross-outcomes} illustrates, the presence of complete supporting evidence becomes substantially clearer in the representation, while the current task head does not automatically exploit that direction. This separation delineates the role of \MAA---retaining information---from the role of the decoder and objective---using that information through an appropriate decision rule.

\begin{fixedfigure}
\centering
\includegraphics[width=\columnwidth]{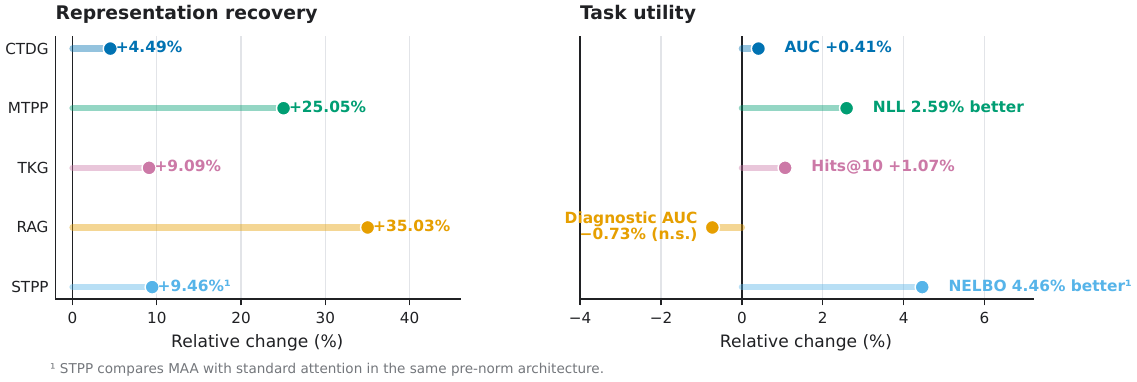}
\captionof{figure}{Relationship between representation accessibility and task utility. Recovery and representative task metrics improve together in CTDG and MTPP. In TKG, ICEWS14 Hits@10 improves while ICEWS18 ranking remains nearly unchanged. In RAG, the retained-passage condition becomes more recoverable without a corresponding improvement in the diagnostic head's AUC.}
\label{fig:cross-outcomes}
\end{fixedfigure}

\subsection{STPP: The Boundary Imposed by Downstream Normalization}
DeepSTPP constructs a nonparametric spatiotemporal intensity through a latent process \cite{zhou2022deepstpp}, whereas Neural STPP conditions on preceding events through an attentive continuous-time flow \cite{chen2021neuralstpp}. For DeepSTPP, we use the numbers of preceding events within radii 0.25, 0.5, and 1.0 of a candidate location as local-count targets. In the original architecture, recovery improves at the final attention output, but the effect disappears immediately after LayerNorm and becomes negative at the predictor-facing $Z$. This trace shows that \MAA creates the signal at the attention boundary and that a subsequent amplitude-removing operation interrupts its propagation.

\begin{figure*}[!t]
\centering
\includegraphics[width=\textwidth]{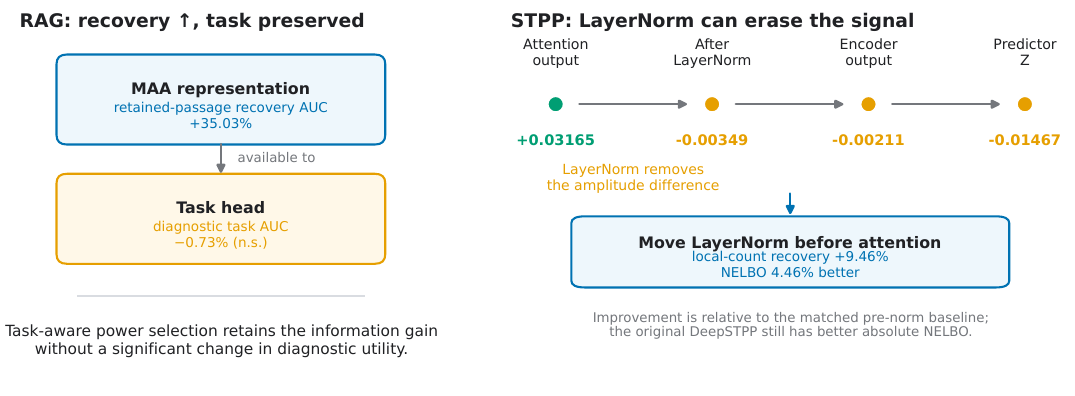}
\caption{Cross-domain boundaries. In RAG, increased information accessibility is separated from task utilization. In STPP, post-attention LayerNorm removes the scale created at the attention output; moving LayerNorm before attention in a matched architecture improves both predictor-facing recovery and NELBO.}
\label{fig:cross-boundary}
\end{figure*}

In a matched pre-norm architecture that places LayerNorm before attention \cite{xiong2020layer}, predictor-facing recovery increases by 9.46\% and NELBO improves by 4.46\%. Its absolute NELBO, however, remains below that of the original DeepSTPP architecture. Figure~\ref{fig:cross-boundary} summarizes this boundary. \MAA retains the effective-support scale at the attention boundary; propagation to the final representation depends on whether subsequent architectural operations preserve that scale.

\section{Mechanism Analysis and Alternative Explanations}
\label{sec:analysis}

This section tests whether the observed improvements are attributable to a training schedule, uniform norm expansion, or explicit cardinality injection.

\subsection{Power or Schedule?}
\begin{fixedtable}
\centering
\captionof{table}{Fixed-power sensitivity. Values are averaged over four representative conditions and characterize the effect of mildly relaxing exact repetition invariance at $p=1$.}
\label{tab:power}
\small
\resizebox{\columnwidth}{!}{\begin{tabular}{@{}rrrrrrr@{}}
\toprule
$p$ & AUC & $\Delta$AUC & Shared linear $R^2$ & $\Delta R^2$ & Context margin & $\Delta$margin \\
\midrule
1.00 & 0.84559 & -- & 0.38277 & -- & 0.48446 & -- \\
1.02 & 0.84809 & +0.30\% & 0.40403 & +5.56\% & 0.51870 & +7.07\% \\
1.05 & 0.84891 & +0.39\% & 0.40626 & +6.14\% & 0.52151 & +7.65\% \\
1.10 & 0.84935 & +0.45\% & 0.40891 & +6.83\% & 0.52465 & +8.29\% \\
1.20 & \best{0.84992} & \best{+0.51\%} & \best{0.41288} & \best{+7.87\%} & 0.53985 & +11.43\% \\
2.00 & 0.84553 & -0.01\% & 0.40940 & +6.96\% & \best{0.60774} & \best{+25.45\%} \\
\bottomrule
\end{tabular}}
\end{fixedtable}

\begin{figure*}[!t]
\centering
\includegraphics[width=.92\textwidth]{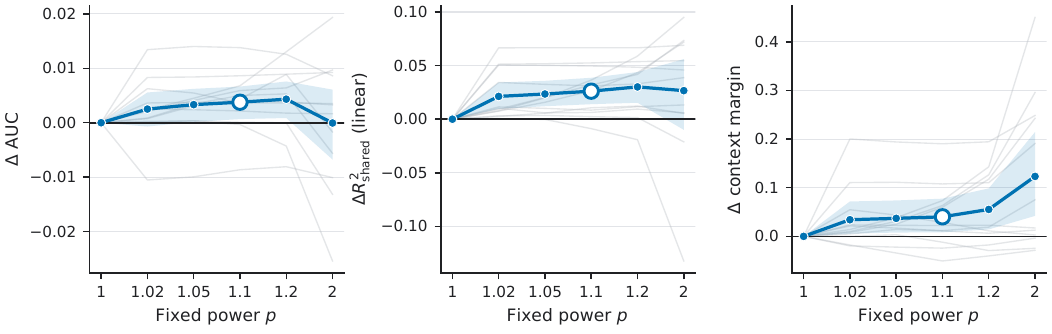}
\caption{Fixed-power sensitivity. Mild normalization relaxation improves recovery and context margin while maintaining or improving AUC. Larger $p$ further increases sensitivity but can weaken the balance with task utility.}
\label{fig:power}
\end{figure*}

Our initial hypothesis assigned importance to a schedule that began with a large $p$ and strengthened invariance later in training. The differences among fixed $p=1.1$, forward $2\!\rightarrow\!1.1$, and reverse $1.1\!\rightarrow\!2$ are small. By contrast, the range 1.02--1.2, which mildly departs from $p=1$, consistently improves AUC, shared linear recovery, and context margin in Table~\ref{tab:power}. Figure~\ref{fig:power} shows both the mean changes and variation across conditions. At $p=2$, context margin increases substantially while the AUC gain disappears. The principal effect therefore arises from mildly relaxing exact $L_1$ repetition invariance rather than from the order of training.

\subsection{Uniform Scaling or an Input-Dependent Effect?}
\begin{figure*}[!t]
\centering
\includegraphics[width=.94\textwidth]{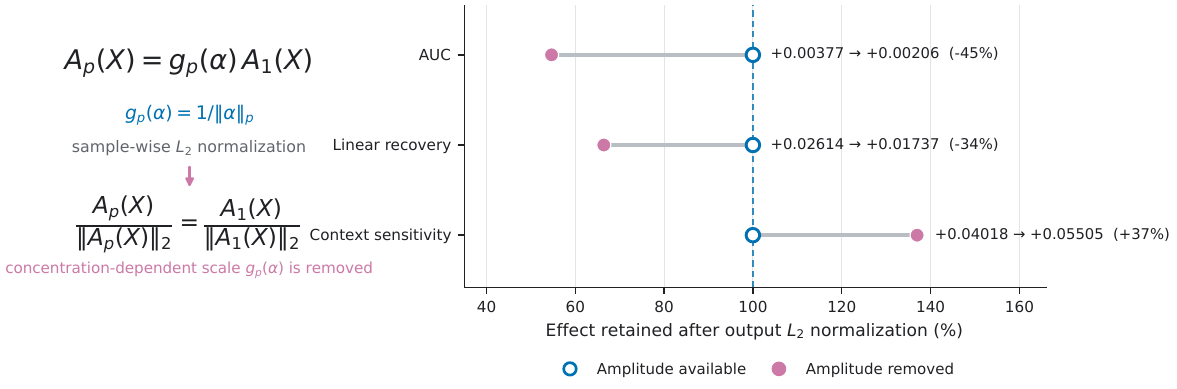}
\caption{Mechanism validation. Removing the $L_2$ norm of the \MAA output attenuates, but does not eliminate, the effect. Input-dependent amplitude is a principal signal path, accompanied by a smaller change in the learned hidden direction.}
\label{fig:mechanism}
\end{figure*}

Equation~\eqref{eq:gate} shows that \MAA multiplies the standard weighted direction by an input-dependent amplitude. To test the alternative explanation that a larger representation norm merely facilitates optimization, we use a matched control that $L_2$-normalizes the aggregation output. Removing the norm attenuates but does not eliminate the gains in AUC, shared linear recovery, and beyond-posthoc recovery. Figure~\ref{fig:mechanism} indicates that amplitude is a principal transmission path and that training also induces some directional change in the representation. The relevant mechanism is therefore the input-varying $g_p(\alpha)$ and its associated optimization trajectory, rather than a uniform scale factor.

\subsection{Comparison with Explicit Cardinality}
\begin{fixedtable}
\centering
\captionof{table}{Matched comparison of MAA with CPA, an explicit cardinality control. Each cell reports a change from baseline; for CPA, the better result between its additive and scaled variants is shown for each endpoint.}
\label{tab:cpa}
\small
\setlength{\tabcolsep}{4pt}
\resizebox{\columnwidth}{!}{\begin{tabular}{@{}llrrr@{}}
\toprule
Representative setting & Endpoint & Best CPA $\Delta$ & MAA $\Delta$ & MAA--CPA \\
\midrule
CRAFT--LastFM & AUC & -0.00258 & \best{+0.00293} & \best{+0.00550} \\
 & Shared linear $R^2$ & -0.00684 & \best{+0.01527} & \best{+0.02211} \\
TGAT--MOOC & AUC & -0.01582 & \best{+0.00159} & \best{+0.01741} \\
 & Shared linear $R^2$ & \best{+0.12934} & +0.03166 & -0.09768 \\
AttNHP--StackOverflow & NLL reduction & -0.02067 & \best{+0.04456} & \best{+0.06523} \\
 & Count $R^2$ & -0.02380 & \best{+0.00213} & \best{+0.02594} \\
THP--StackOverflow & NLL reduction & \best{+0.24629} & +0.18438 & -0.06190 \\
 & Count $R^2$ & +0.08140 & \best{+0.08650} & +0.00511 \\
xERTE--ICEWS14 & MRR & -0.00216 & -0.00341 & -0.00125 \\
 & Pair-recurrence $R^2$ & -0.06946 & \best{+0.00128} & \best{+0.07074} \\
xERTE--ICEWS18 & MRR & -0.01280 & \best{-0.00047} & \best{+0.01233} \\
 & Pair-recurrence $R^2$ & -0.07276 & \best{+0.00579} & \best{+0.07855} \\
\bottomrule
\end{tabular}}
\end{fixedtable}

CPA directly injects the number or unweighted sum of inputs and can substantially improve recovery or task performance in particular conditions, as observed for TGAT--MOOC recovery and THP NLL. The same intervention, however, strongly perturbs performance or recovery for CRAFT, AttNHP, and xERTE. \MAA instead reflects the effective support of the learned attention distribution without a separate count, producing a more stable representation--task trade-off. In Table~\ref{tab:cpa}, \MAA outperforms the best CPA variant in recovery for five of six representative cases and in task utility for four of six. This comparison empirically distinguishes the retention of attention-weighted support from the direct reinjection of counts.

\subsection{What Is Learned, and Where Is It Useful?}
The information made more accessible by \MAA has domain-specific names but a common structure: the extent of evidence accumulated across multiple inputs. In CTDG, PA and pair frequency describe pair- and node-level accumulation induced by repeated edge formation. Candidate-mark count in MTPP is a direct input to the intensity of the next event, and pair recurrence in TKG provides query-specific support for a candidate tail. The retained-passage condition in RAG indicates whether all distinct evidence required for an answer is present, while local count in STPP represents conditional event density around a candidate location.

This information supports three uses. First, when the original task head exploits it, the improvement appears in metrics such as AUC, NLL, and ranking quality. Second, linear accessibility from the same $Z$ enables model auditing, transfer heads, calibration, and downstream decisions even when the original head does not immediately convert it into task performance. Third, tracing where accessibility disappears reveals architectural information bottlenecks. RAG most clearly separates accessibility from direct utilization, while STPP exposes the downstream normalization boundary.

\subsection{Limitations}
\MAA is a general principle for improving representation informativeness rather than a universal task-performance optimizer. Information reaches the predictor-facing $Z$ when the path following attention preserves amplitude, and the task head converts it into utility when its objective is aligned with that information. Recovery targets likewise follow the semantically meaningful evidence unit of each domain. Because the cross-domain samples are smaller than the CTDG study, we interpret them through effect sizes, replication counts, and task performance together. Broader validation across architectures and large-scale real-world tasks remains valuable. Although power selection is separated from evaluation, the validation protocols are not identical across domains; a transferable selection rule based on the distribution of $g_p(\alpha)$ is a promising direction.

\section{Conclusion}
\label{sec:conclusion}

Standard attention is a weighted average and can therefore map repetitions of the same evidence pattern to an identical aggregate. We introduced Mass-Aware Attention, which turns this implicit $L_1$ repetition invariance into an explicit control axis within an $L_p$ family. The method changes a single denominator and adds no supervision, count feature, or parameter.

In CTDG, \MAA increases the linear accessibility of graph statistics from the same hidden representation, improving preferential-attachment recovery in all 12 model--dataset cells. Future-link AUC also increases in 11 of 12 cells. Candidate-mark counts and pair recurrence become more recoverable in MTPP and TKG, while RAG and STPP reveal how information retention interacts with task heads and downstream normalization.

Together, these results connect representation gaps observed under different names to a shared blind spot of weighted averaging and show that a minimal normalization principle can make accumulated evidence more accessible. \MAA advances attention design beyond judging learning solely through predictive performance by making the evidence transmitted to the final representation an explicit design consideration.

\section*{Declaration of competing interest}
The authors declare that they have no known competing financial interests or
personal relationships that could have appeared to influence the work reported
in this paper.

\section*{Acknowledgements}
The authors have no acknowledgements to declare.

\section*{Funding}
This research received no specific grant from any funding agency in the public,
commercial, or not-for-profit sectors.

\section*{Data availability}

All datasets used in this study are publicly available. The exact data sources are listed in Table~\ref{tab:data_availability}.

\begin{fixedtable}
\centering
\captionof{table}{Public dataset sources used in this study.}
\label{tab:data_availability}
\small
\setlength{\tabcolsep}{4pt}
\renewcommand{\arraystretch}{1.15}
\scriptsize
\begin{tabularx}{\columnwidth}{@{}llX@{}}
\toprule
Domain & Dataset & Source \\
\midrule
CTDG & Wikipedia & \url{https://snap.stanford.edu/jodie/wikipedia.csv} \\
CTDG & MOOC & \url{https://snap.stanford.edu/jodie/mooc.csv} \\
CTDG & LastFM & \url{https://snap.stanford.edu/jodie/lastfm.csv} \\
MTPP & StackOverflow & HuggingFace: \url{https://huggingface.co/datasets/easytpp/stackoverflow} \\
MTPP & Retweets & HuggingFace: \url{https://huggingface.co/datasets/easytpp/retweet} \\
STPP & Earthquake & GitHub: \url{https://github.com/ss15859/EarthquakeNPP} \\
STPP & Gowalla & \url{https://snap.stanford.edu/data/loc-gowalla_totalCheckins.txt.gz} \\
TKG & ICEWS14 & GitHub RE-Net data: \url{https://github.com/INK-USC/RE-Net} \\
TKG & ICEWS18 & GitHub RE-Net data: \url{https://github.com/INK-USC/RE-Net} \\
RAG & HotpotQA & HuggingFace: \url{https://huggingface.co/datasets/hotpotqa/hotpot_qa} \\
\bottomrule
\end{tabularx}
\end{fixedtable}

\section*{Code availability}

The implementation, experiment configurations, aggregation scripts, and plotting scripts used in this study are available at \url{https://github.com/SnowyPainter/maa-public}.

\bibliographystyle{IEEEtran}
\bibliography{references}

\appendices
\section{Definitions and Application Details}
\label{app:details}

This appendix specifies the temporal scope of the recovery targets and the attention boundaries at which \MAA is applied.

\subsection{Recovery Targets}
All recovery targets are computed solely from observations preceding the query time. Table~\ref{tab:target-definitions} reports the exact formulas, transformations, fallback values, and coverage used in the implementation. Homophily is the continuous Jaccard similarity between endpoint neighborhoods, not a binary community label. Explicit fallback values make every target, including recency and periodicity, numerically defined for every query.

All eight CTDG targets are evaluated by $R^2$ on the same chronological held-out split, and broad recovery is their arithmetic mean. Evidence support in the table is the proportion of queries with observed evidence for the statistic rather than its fallback value; it is not used as an evaluation mask. Only the binary RAG evidence condition is evaluated separately with AUC.

\begin{table*}[!t]
\centering
\caption{Exact definitions of the CTDG recovery targets. Every value is computed from observations preceding query time $t$ and is numerically defined for every query. Support reports the dataset-wise proportion of queries with observed evidence rather than a fallback value.}
\label{tab:target-definitions}
\scriptsize
\setlength{\tabcolsep}{3.5pt}
\begin{tabularx}{\textwidth}{@{}lXXlll@{}}
\toprule
Target & Definition & Transformation and fallback & Metric & Query coverage & Evidence support \\
\midrule
Pair frequency
& Cardinality of the preceding pair-event set $T_{uv}$
& $\log(1+|T_{uv}|)$; 0 for an unseen pair
& $R^2$ & 100\% & 28.05--42.82\% \\
Recency
& $t-\max T_{uv}$
& $\log(1+\cdot)$; source-history age for an unseen pair, or 0 if the source also has no history
& $R^2$ & 100\% & 28.05--42.82\% \\
Persistence
& Fraction of $B$ bins in the source-history span occupied by pair events
& $B=\min(10,\max(2,\lfloor\sqrt{|T_u|+1}\rfloor))$; 0 with fewer than two pair events
& $R^2$ & 100\% & 16.21--39.32\% \\
Periodicity
& Regularity of pair-event intervals $G=\Delta T_{uv}$
& $1/(1+\operatorname{std}(G)/(\operatorname{mean}(G)+10^{-12}))$; 0 with fewer than three events or a zero mean interval
& $R^2$ & 100\% & 9.76--36.86\% \\
Granularity
& Median positive interval between source events
& $\log(1+\operatorname{median}(\Delta T_u^+))$; 0 if no positive interval exists
& $R^2$ & 100\% & 89.18--99.22\% \\
Density
& Fraction of observed source neighbors $|N(u)|/(|V|-1)$
& No additional transformation
& $R^2$ & 100\% & 100\% \\
Homophily
& Endpoint-neighborhood Jaccard similarity $|N(u)\cap N(v)|/|N(u)\cup N(v)|$
& 0 if the union is empty
& $R^2$ & 100\% & 97.87--99.97\% \\
Preferential attachment
& Endpoint-degree product $d_ud_v$ immediately before the query
& $\log(1+d_ud_v)$
& $R^2$ & 100\% & 100\% \\
\bottomrule
\end{tabularx}
\end{table*}

\subsection{Attention Boundaries}
Within CTDG, \MAA is applied only to the temporal aggregation that forms the predictor-facing representation: CRAFT's target-aware cross-attention, TGAT's query-to-neighbor attention, DyGFormer's final query attention over encoded patches, and GraphMixer's final query attention over mixed tokens. Each model's internal Transformer, token mixer, feature encoder, and predictor remain unchanged.

The same rule governs the other domains. We first identify an attention boundary of the form $\sum_i\alpha_i v_i$ that forms a predictor-facing representation, retain the score and value computations, and replace only the normalization denominator from $L_1$ to $L_p$. This rule standardizes the scope of modification across models and ensures that recovery changes do not arise from additional hidden dimensions or features.

\balance
\end{document}